\documentclass[utf8]{FrontiersinHarvard} 

\usepackage{url,hyperref,lineno,microtype,subcaption}
\usepackage[onehalfspacing]{setspace}
\usepackage{bm}
\usepackage{amsmath}
\def\bfm#1{{\bf #1}}

\def\bfs#1{\mbox{\boldmath{$ #1 $}}}

\usepackage{xcolor, soul}
\usepackage{wrapfig}
\usepackage{mathtools}
\usepackage{multirow}
\DeclarePairedDelimiter{\norm}{\lVert}{\rVert}
\definecolor{fgwhite}{rgb}{1,1,1}     
\definecolor{fgred}{rgb}{0.8,0,0}     
\definecolor{fgorange}{rgb}{0.93,0.53,0.18}     
\definecolor{fgbrown}{rgb}{0.69,0.35,0}     
\definecolor{fgpurple}{rgb}{0.55,0.1,0.6}     
\definecolor{fggreen}{rgb}{0,0.5,0}     
\definecolor{bggreen}{rgb}{0.8,1,0.8}     
\definecolor{fgblue}{rgb}{0,0,0.7}     
\definecolor{bgblue}{rgb}{0.9,0.9,1}     
\definecolor{fgclay}{rgb}{0.51,0.25,0.04}  
\definecolor{paleyellow}{rgb}{0.9,0.9,0.06}  
\definecolor{forIan}{rgb}{0.0,0.0,0.5}

\usepackage{array}
\newcolumntype{L}{>{\centering\arraybackslash}m{4.5cm}}
\newcolumntype{R}{>{\centering\arraybackslash}m{2.5cm}}

\def\keyFont{\fontsize{8}{11}\helveticabold }
\def\firstAuthorLast{Deshpande {et~al.}} 
\def\Authors{Saurabh Deshpande\,$^{1}$, Ra\'ul I. Sosa\,$^{2}$, St\'ephane P.A. Bordas\,$^{1,*}$ and Jakub Lengiewicz\,$^{1,3}$}

\def\Address{$^{1}$Department of Engineering; Faculty of Science, Technology and Medicine; University of Luxembourg \\
$^{2}$Department of Physics and Materials Science; Faculty of Science, Technology and Medicine; University of Luxembourg \\
$^{3}$Institute of Fundamental Technological Research, Polish Academy of Sciences  }
\def\corrAuthor{St\'ephane P.A. Bordas}
\def\corrEmail{stephane.bordas@alum.northwestern.edu}

\begin{document}
\onecolumn
\firstpage{1}

\title[Convolution, aggregation \& attention based DNNs]{Convolution, aggregation and attention based deep neural networks for accelerating simulations in mechanics}

\author[\firstAuthorLast ]{\Authors} 
\address{} 
\correspondance{} 

\extraAuth{}%

\maketitle

\begin{abstract}

\section{}

Deep learning surrogate models are being increasingly used in accelerating scientific simulations as a replacement for costly conventional numerical techniques. However, their use remains a significant challenge when dealing with real-world complex examples. In this work, we demonstrate three types of neural network architectures for efficient learning of highly non-linear deformations of solid bodies. The first two architectures are based on the recently proposed CNN U-NET and MAgNET (graph U-NET) frameworks which have shown promising performance for learning on mesh-based data. The third architecture is Perceiver IO, a very recent architecture that belongs to the family of attention-based neural networks--a class that has revolutionised diverse engineering fields and is still unexplored in computational mechanics. We study and compare the performance of all three networks on two benchmark examples, and show their capabilities to accurately predict the non-linear mechanical responses of soft bodies.

\tiny
 \keyFont{ \section{Keywords:} Surrogate Modeling, Deep Learning, CNN U-NET, Graph U-Net, Perceiver IO, Finite Element Method} 
\end{abstract}

\section{Introduction}

The ability to make fast or real-time predictions of the response of physical systems is essential for a variety of engineering applications. Notable examples of this can be found in the field of robotics \citep{rus2015design, choi2021use} and medical simulations \citep{courtecuisse2014real,cotin,bui,MAZIER2021110645}, which have the potential to advance personalized medicine and improve computer-assisted and robotic surgery~\citep{chen2020deep,Dennler2021}. In computational physics and chemistry, fast and accurate predictions are fundamental for studying complex systems, such as those arising in biology and materials science~\citep{Friesner2005}, or in drug discovery~\citep{DeVivo2016}. In many cases, the necessary accuracy of these predictions requires complex models that can be expressed through partial differential equations and solved numerically using methods such as the finite element method (FEM) at continuum scales or specialized \emph{ab initio} approaches at the atomic or quantum scales. However, these high-fidelity computational models are often too slow for real-time or practical purposes, and therefore approximate or surrogate models must be developed to achieve the necessary speed-ups.

At the same time, the 21st century has seen an explosion of measurement data, much of which is available as public datasets in various scientific domains, including structural mechanics, material science, and meteorology \citep{zakutayev2018open,ELOUNEG2022107835,GHOLAMALIZADEH2022107009}. The availability of this data, combined with the rapid growth in computational resources, has led to the increasing importance of machine learning (ML) techniques \citep{butler2018machine,bock2019review,schleder2019dft} for solving forward and inverse engineering problems. This includes surrogate and data-driven approaches that aim to enable modeling \citep{ma12162574}, accelerate computationally costly direct numerical simulations \citep{RuppEtAl2012,wirtz2015surrogate, CAPUANO2019363,weerasuriya2021gaussian}, and even discover new material laws \citep{LIU2017159,euclid}. The increasing use of ML in engineering and other fields has also spurred the development of various methods and algorithms for improving the accuracy and efficiency of these techniques.

Within the class of machine learning methods for surrogate and data-driven modeling, deep learning (DL) approaches have seen great success due to their ability to efficiently extract complex relationships present in the underlying data. DL models have been successfully employed for a range of tasks in diverse fields such as computational physics\&chemistry, material science, computational mechanics, computer vision, natural language processing, and many others \citep{ schutt2017quantum,jha2018elemnet,voulodimos2018deep,OISHI2017327,schmidt2019recent,brown2020language,choudhary2022recent}. In computational chemistry, machine learning force fields (MLFFs), see~\citep{Unke2021}, have seen great success in recent years for accelerating costly \emph{ab initio} simulations. For instance, Deep Tensor Neural Network (DTNN) \citep{schutt2017quantum} and SchNet \citep{NIPS2017_303ed4c6} models have been shown to accurately predict forces in a variety of molecules and could be used in applications such as protein folding and material design.
Similarly, computational mechanics has witnessed an increasing use of DL surrogate models as a replacement for costly direct numerical simulations \citep{mianroodi2021teaching,ABUEIDDA2021102852}. What is common to all the above-mentioned cases is that deep learning techniques rely on deep artificial neural networks (deep ANNs, or DNNs), which must be trained on a sufficiently large amount of data. While this training process is computationally costly, once trained, the predictions of DL models are extremely efficient.

Obtaining necessary amount of training data is often difficult when it originates from physical experiments. 
This can be due to multiple factors, such as high costs, risks \& difficulties associated with the experiments, or data privacy clauses. 
There are two possible approaches to deal with the scarcity of experimental data. The first approach relies on enhancing the DL model with the information on underlying physics -- an approach popularly termed as Physics Informed Neural Networks (PINN) \citep{earlyPINN,MAO2020112789,SAMANIEGO2020112790,alban}. The second approach includes the underlying physics implicitly, through high-fidelity simulations done \emph{in silico} to provide the necessary amount of synthetically generated data, which has shown to be useful in various applications \citep{le2017using,kim2022how,Pfeiffer2019,aydin2019general,soumi}. In this work, we will follow the latter approach and will focus on DL surrogate models that are trained on synthetically generated data from finite element simulations in non-linear elasticity.

One of the most important aspects that will be studied in this work is the architecture of deep neural networks. The majority of DL approaches that are present in the literature are based on fully connected networks, which can be inefficient and prone to overfitting when applied to high-dimensional inputs. If such large inputs are structured, they fall under the umbrella of geometric deep learning (GDL) \citep{bronstein2021geometric}, a concept that has gained increasing interest in recent years. In this work, we will compare three architectures that can efficiently handle high-dimensional structured inputs: convolutional neural networks (CNNs), graph neural networks (GNNs), and attention-based networks.

Convolutional neural networks (CNNs) are known to outperform traditional fully-connected ANNs, and this has been demonstrated in various domains, including physics-based simulations \citep{10.1145/2939672.2939738,vasilis,doi:10.1063/5.0077723,DESHPANDE2022115307}. CNNs work on the principle of parameter sharing and local convolution operations, which enables efficient training on large inputs. Their disadvantage is that the inputs/outputs of CNNs are restricted to grid inputs, such as images, videos, or structured FE meshes. However, CNNs have found their successors, the graph neural networks (GNNs), that can work with any structure of inputs/outputs.

Graph-based approaches leverage the topological information of the input to perform local operations in the respective neighborhood only, and can learn efficiently on generally structured data. Recently, GDL methods have shown promising performance for their applications as well in the field of  mechanics, \citep{battaglia2018relational,VLASSIS2020113299,pfaff2021learning,vasilisgraph,sebstornich}. More recently, \citep{magnet} proposed MAgNET, a novel graph U-Net framework for efficiently learning on mesh-based data. In this work we utilise it to accurately predict non-linear deformations of solids.

Attention-based approaches, similar to human cognitive attention, work by allowing the DL model to focus on certain parts of the input data that are relevant to the task at hand. This is done through a fully trainable process that, without the need to introduce topological information or enforce structural restrictions, allows the neural network to extract dependencies from throughout the whole input domain. This type of approach has led to significant strides in a wide range of areas, starting from computer vision \citep{xu2015show} to natural language processing \citep{devlin2018bert,baevski2020wav2vec}, as well as becoming the basic building block of the Transformer architecture \citep{vaswani2017attention}. Recently the Perceiver~IO~\citep{jaegle2022perceiver}, a new type of architecture that builds upon Transformers, has been proposed as a general-purpose model that can handle data from arbitrary settings. Since Perceiver IO has been shown to achieve several state-of-the-art results without the need for problem-specific architecture engineering, we will compare its performance on non-linear deformation prediction of solids based on mesh data against the previously discussed models.  

To summarise, in this work we will compare three DNN architectures: two architectures presented in our earlier works, i.e., CNN U-Net framework~\citep{DESHPANDE2022115307}, and MAgNET framework~\citep{magnet}, as well as the attention-based architecture, Perceiver~IO \citep{jaegle2022perceiver}, which has not been explored for its applications in mechanics yet. We show the capabilities of three frameworks by learning on non-linear FEM datasets and by cross-comparing their performance. In Section~\ref{sec: methodlogy}, we will introduce the three DNN architectures, in Section~\ref{sec: results}, we will study their performance, and in Section~\ref{sec: conclusion} we will summarize the results and discuss future directions.

\section{Method}\label{sec: methodlogy}

\begin{figure}[t!]
\begin{center}
\includegraphics[width=1\textwidth]{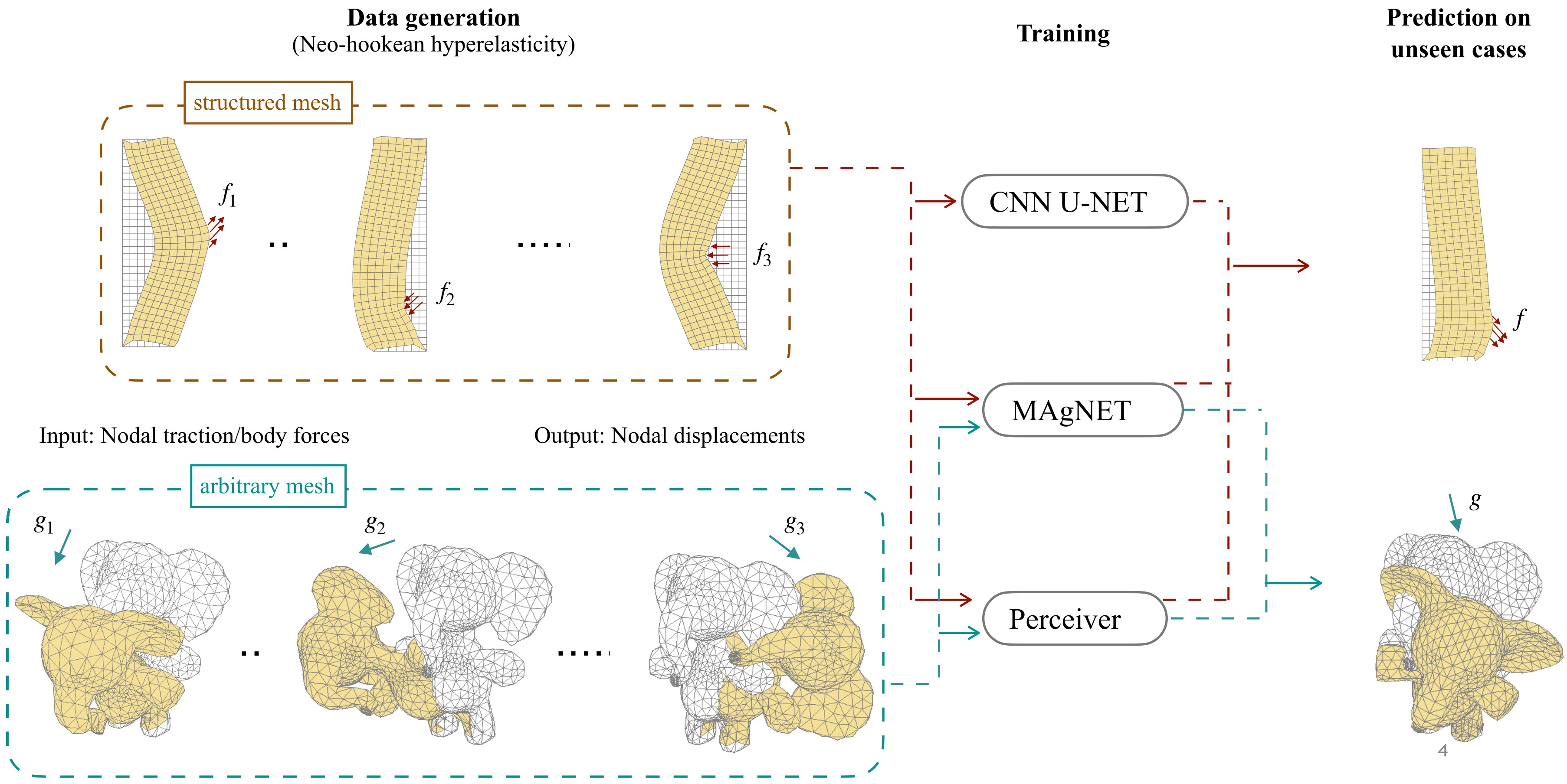}
\end{center}
\caption{Outline of the neural network surrogate frameworks for predicting body deformations. (Left) Training datasets for structured and arbitrary mesh cases are generated by using a non-linear FEM solver. (Middle) Proposed neural network frameworks are trained on these datasets. For structured mesh case, all NN frameworks are used while for arbitrary unstructured meshes only MAgNET and Perceiver~IO networks are used. (Right) Trained networks are then used as surrogate models to predict the deformation of bodies under unseen forces.}\label{fig: outline}
\end{figure}

As previously mentioned in the introduction, in this paper we propose three types of deep neural network (DNN) frameworks that can be used as surrogate models to replace computationally expensive non-linear FEM solvers. The proposed DNN frameworks are trained on force-displacement FEM datasets that are given in the mesh format. Once trained, these surrogate DNN models are able to quickly and accurately simulate the mechanical responses of bodies subjected to external forces. The outline of study pursued in this paper is shown in Figure~\ref{fig: outline}.

Any input mesh can be categorised into either a structured mesh or an arbitrary unstructured mesh. In this work we introduce three types of DNN network architectures. The CNN U-Net network can only be (straightforwardly) used for structured meshes, while the MAgNET and Perceiver~IO networks are more general and are capable of handling arbitrary mesh inputs. All these frameworks are discussed in detail in the following subsections.

\subsection{DNN frameworks for predicting mechanical deformations}

Below we introduce three different types of neural network architectures which can efficiently predict non-linear deformations of bodies subjected to external traction and body forces. All the proposed DNN frameworks directly operate on the finite element mesh data thereby making it very convenient to be used as surrogate models in place of conventional FEM solver. The first two i.e. CNN U-Net and MAgNET belong to the family of U-Net architecture while Perceiver IO is based on Transformer-type attention, see Table~\ref{tab:frameworks}.

\bgroup
\def\arraystretch{1.4}
\begin{table}[h!]
    \centering
    \begin{tabular}{c|c|c}
        Framework & Type & Supported mesh  \\
        \hline
        CNN & U-Net (convolution operation) & structured    \\
        MAgNET  & Graph U-NET (MAg operation)  &arbitrary   \\ 
        Perceiver~IO & Transformer (attention mechanism) & arbitrary   
    \end{tabular}
    \captionsetup{justification=centering}
    \caption{Properties of deep neural network architectures studied in this work.}
    \label{tab:frameworks}
\end{table}
\egroup

\subsubsection{CNN U-Net}
\begin{figure}[h!]
\begin{center}
\includegraphics[width=0.8\textwidth]{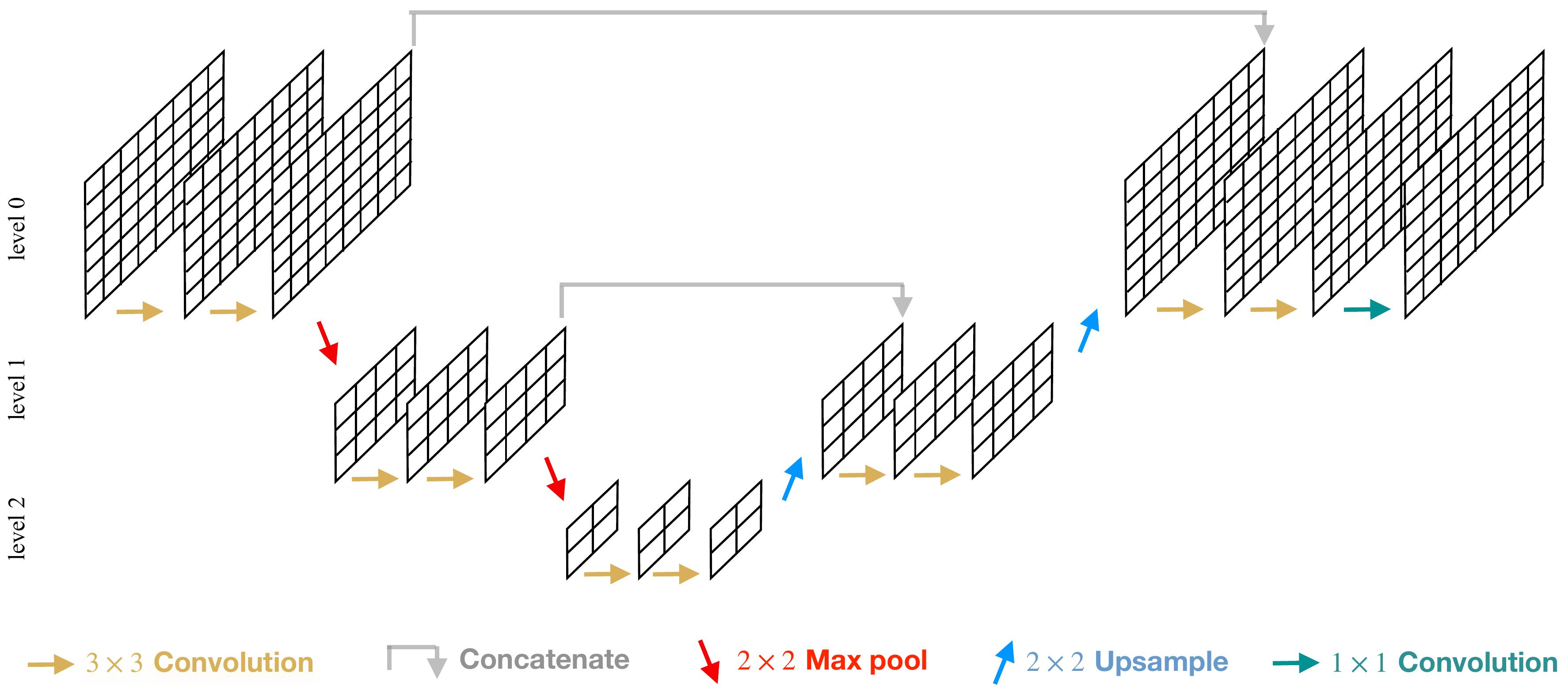}
\end{center}
\caption{Schematic of CNN architecture used for generic structured 2D mesh inputs.}\label{fig: CNN schematic}
\end{figure}

CNNs were originally proposed for performing classification and regression tasks on image, video like data but lately are even being used for generic inputs such as mesh data which is crucial to many scientific applications. In particular, U-Net like architectures have shown great potential in learning on large-scale inputs and lately have been successfully used for simulating mechanical responses of materials as well \citep{mianroodi2021teaching,DESHPANDE2022115307}. The name U-Net comes from the particular U-shaped architecture which involves a series of convolutional and pooling operations. Convolutional layers are responsible for non-linear transformations whereas pooling enables learning through low-fidelity representation thus making the network capable of learning on high-dimensional inputs. Experiments presented in this work are carried out by using the CNN U-Net framework proposed by \citep{DESHPANDE2022115307}, see Figure~\ref{fig: CNN schematic}.

One major limitation of the CNN is that it cannot straightforwardly accommodate unstructured mesh inputs. To overcome this issue, the simplest approach embeds a structured grid on unstructured meshes with a naive mapping between unstructured and grid node values \citep{U-Mesh}. While more sophisticated approaches are proposed to make unstructured meshes compatible to be used with CNN framework \citep{immerserdboundary}. However, they perform poorly on complicated geometries and come with an associated preprocessing cost; they are not considered in the scope of this work. 

\subsubsection{MAgNET}

\begin{figure}[h!]
\begin{center}
\includegraphics[width=0.9\textwidth]{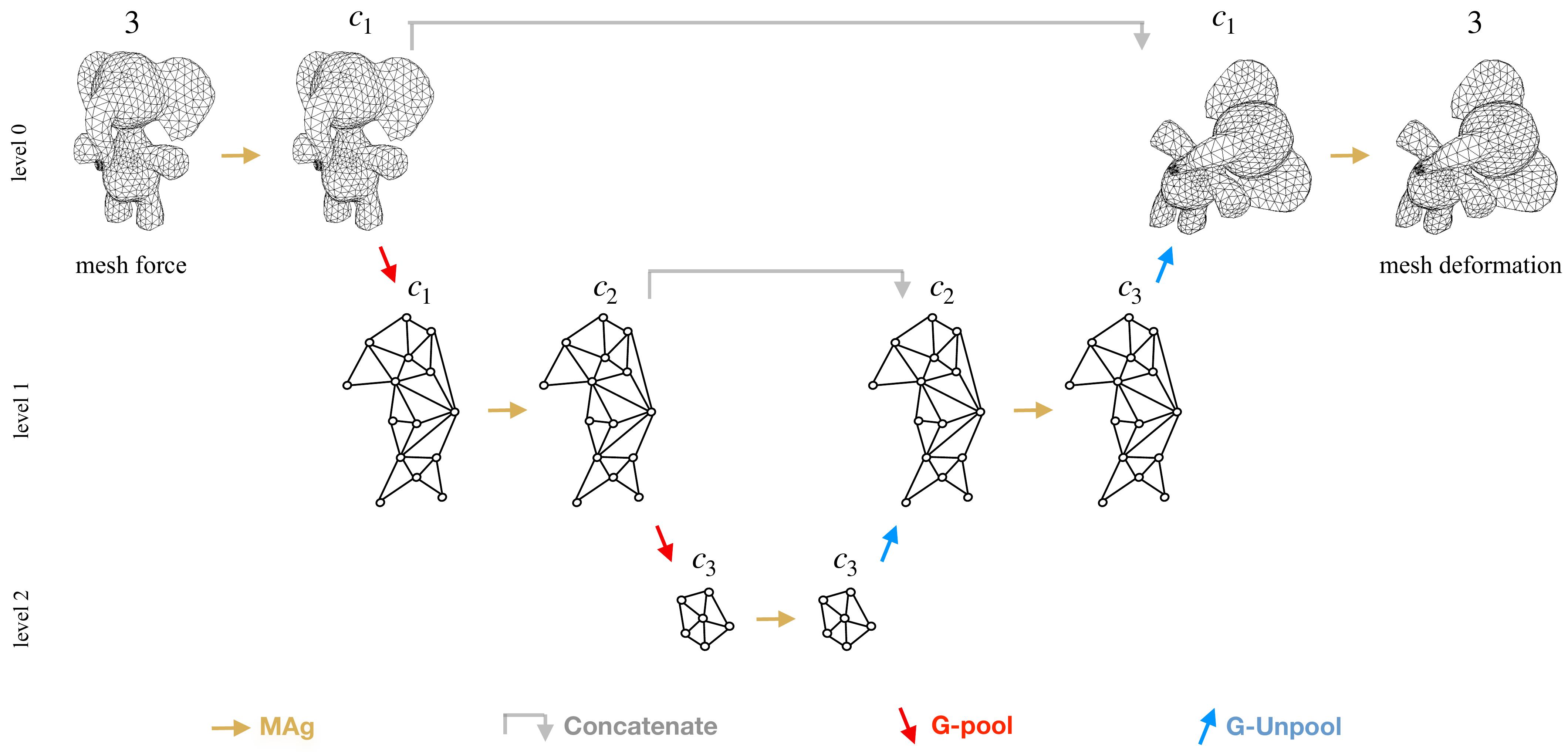}
\end{center}
\caption{Schematic of the MAgNET architecture used in this work. It takes external forces on arbitrary mesh as an input to gives mesh displacements as output.   }\label{fig: magnet_schematic}
\end{figure}

In an attempt to generalise CNN to arbitrary unstructured meshes, very recently \citep{magnet} proposed the MAgNET framework, see Figure~\ref{fig: magnet_schematic}. MAgNET architecture belongs to the family of graph U-Net architectures and it is proposed for efficient learning on mesh structured data. MAgNET directly accepts arbitrary mesh inputs (such as forces/stresses/displacements of nodes in the mesh) values thus making it very convenient to be used with existing numerical solvers. 

MAgNET relies on the so-called MAg layer (Multichannel Aggregation layer) which is capable of learning nonlinear transformations between input and output data existing in the mesh format. MAg extends the concept of local operations in convolution layers to arbitrary mesh inputs by performing aggregation of nodal feature values in the respective neighborhood nodes only. It leverages the topology of inputs and performs learnable local aggregations with heterogeneous window sizes as opposed to the fixed-size window in the case of CNN. While its graph pooling/unpooling layers enable efficient learning on large-dimension inputs through reduced graph representation.

\subsubsection{Perceiver IO}

\begin{figure}[h!]
\begin{center}
\includegraphics[width=0.9\textwidth]{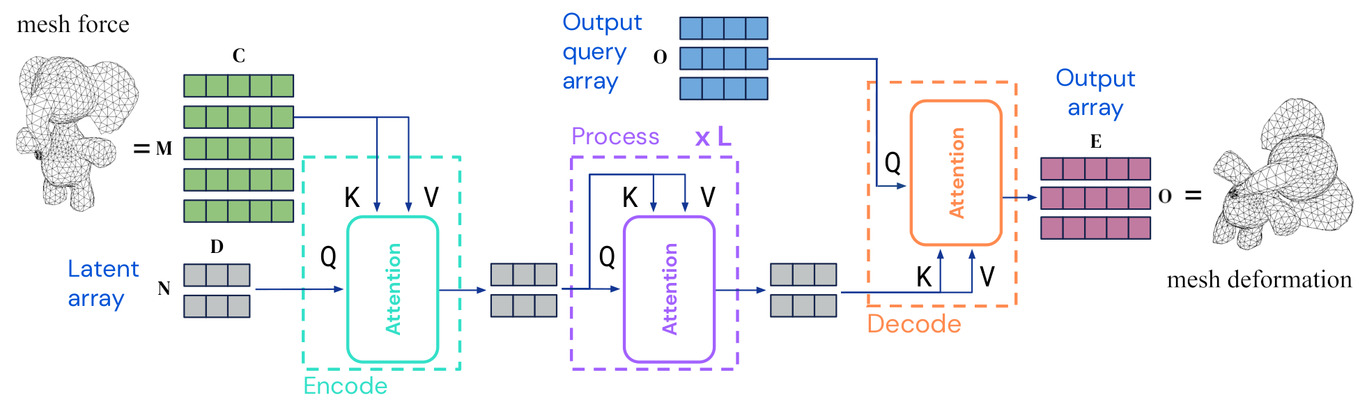}
\end{center}
\caption{Schematic of the Perceiver IO architecture, \citep{jaegle2022perceiver}, used for external forces on arbitrary mesh as inputs and mesh displacement as outputs.}\label{fig: Perceiver schematic}
\end{figure}

The Perceiver IO architecture, \citep{jaegle2022perceiver}, was developed with the goal of achieving a DL scheme that can easily integrate and transform arbitrary information for arbitrary tasks. This architecture employs an attention encoder that maps inputs from a wide range of modalities to a fixed-size latent space using cross-attention, this latent space is then further processed using self-attention as an usual Transformer and decoded into the output domain via cross-attention, see Figure~\ref{fig: Perceiver schematic}. This process allows the network to scale to large and multi-modal data since it decouples the bulk of the network’s processing from the size and modality-specific details of the input. In this work, we will leverage this property and use Perceiver IO to learn non-linear deformations on unstructured meshes without adding any information or restrictions about how to treat the underlying data structure. During training, Perceiver IO automatically learns the important dependencies that exist in the input domain, composed of arbitrarily unstructured mesh data, and transforms them into the corresponding output which consists of displacement data. 

\subsection{Input/output \& training of DNN surrogate models} \label{sec: inouttraining}

As motivated in Section~\ref{sec: methodlogy}, the proposed DNN frameworks are trained on the force-displacement datasets. Let us denote the neural network in consideration as $h$, it is parameterised by trainable parameters, $\bfs{\theta}$. In all the cases, $h$ accepts external forces, \bfm{f}, on all degrees of freedom (dofs) of mesh as the input. And as an output it predicts displacement vector, \bfm{u}, of all dofs (same size as the input), i.e., $h: \bfs{f}\rightarrow\bfs{u}$. 

In the case of CNN and MagNET, forces associated with X, Y, Z degrees of freedom are kept in different channels. For instance, in the case of CNN U-Net, X, Y directional forces on a 2D quad mesh with  $n_x \times n_y$ nodes are fed to the network as a $2 \times n_x\times n_y$ tensor. For MAgNET, they are provided as a single dimensional tensor of shape $2 \cdot n_x \cdot n_y$, with X and Y directional forces concatenated together (each representing a different channel). On the other hand, Perceiver IO inputs are first kept as a single array of size $2 \cdot n_x \cdot n_y$ to which we apply 1 × 1 convolution kernels to project it to a tensor of shape $2 \cdot n_x \cdot n_y \times 256$, adding $256$ channels. Finally, to this channel dimension we concatenate trainable 1D positional embeddings, thus leaving us with a tensor of shape $2 \cdot n_x \cdot n_y \times 512$ that we will use as an input for the network.

Now, for a given training dataset $\{(\bfm{f}_1,\bfm{u}_1),...,(\bfm{f}_N,\bfm{u}_N)\}$, $h$ is trained by minimizing mean squared error between true and predicted values as to get optimised parameters $\bm{\theta}^{*}$ as:  

\begin{equation}
      \bm{\theta}^{*} = \text{arg}\underset{\bm{\theta}}{\text{min}}~\frac{1}{N}\sum_{i=1}^{N} \norm{h(\bfm{f}_i,\bm{\theta})-\bfm{u}_i}_2^{2}
     \label{eq:lossDeterm}
\end{equation}

Performance of $h$ (i.e. the neural network in consideration) over respective test dataset $\{(\bfm{f}_1,\bfm{u}_1),...,(\bfm{f}_M,\bfm{u}_M)\}$ is measured in terms of mean absolute error which is computed for each example ($e_m$) and for the entire test set ($\Bar{e}$) as follows: 
\begin{equation}
     e_{m} = \frac{1}{\mathcal{F}}\sum_{i=1}^{\mathcal{F}}{|h(\bfm{f}_m)^{i} - \bfm{u}_m^{i}|}, \qquad
     \Bar{e} =\frac{1}{M} \sum_{m=1}^{M} e_{m}, \qquad
     \sigma(e) = \sqrt{\frac{1}{M-1}\sum_{m=1}^{M} \left(e_{m} - \Bar{e} \right)^2},
     \label{eq: test loss}
 \end{equation}
where $\mathcal{F}$ stands for the number of dofs of the mesh. The maximum error over the entire test dataset is defined as
 \begin{equation}
    e_{\text{max}}=\max_{m, i}|h(\bfm{f}_m)^{i} - \bfm{u}_m^{i}|.
 \end{equation}

\section{Results}
\label{sec: results}

We validate proposed neural network frameworks on two examples, representing a 2D and a 3D problem respectively. For the 2D example, a structured mesh is considered so that all frameworks including CNN can be applied to it. Whereas for the 3D example an arbitrary unstructured mesh is considered. 

\subsection{Generation of hyperelastic FEM training data}\label{sec: data generation}

As motivated in the methodology section, training datasets of non-linear displacement solutions are generated by applying random traction and body forces on the given discretisation. The number of cases generated randomly (dataset size) has been chosen large enough to generalise well to unseen arbitrary forces. The dataset is split into the training (95\%) and testing (5\%) part, and the pairs of input force and output displacement solutions from the training dataset are then fed to train different types of neural networks. The proposed neural network frameworks are validated on two examples, both following Neo-Hookean hyperelastic law. To avoid the divergence of the non-linear FEM solver, both traction and body forces are applied in incremental load steps. All the computations are performed using AceFEM framework \citep{acegen}.  

\begin{figure}[h!]
\begin{center}
\includegraphics[width=0.9\textwidth]{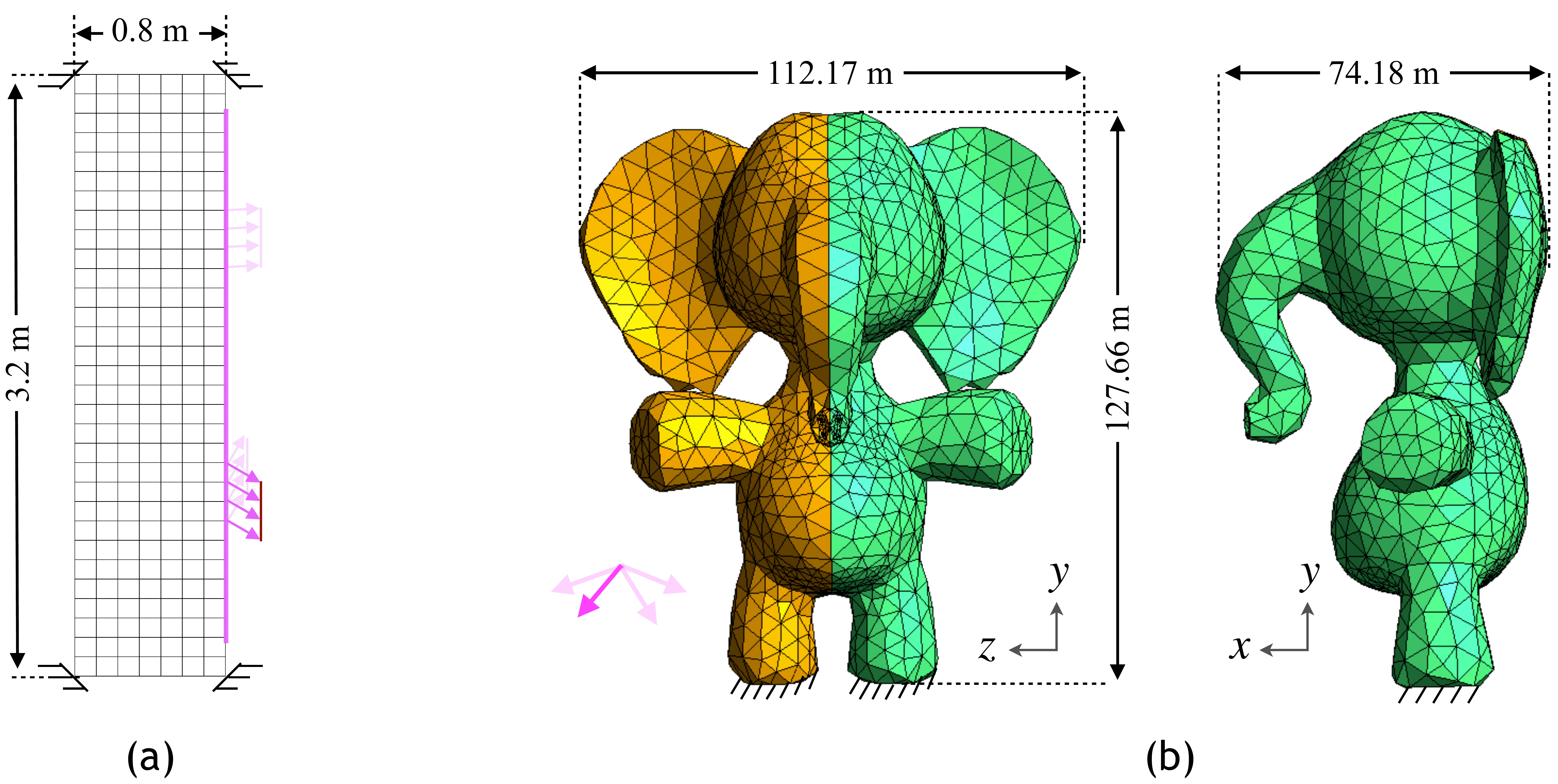}
\end{center}
\caption{Schematics of dataset generation for \textbf{(a)} 2D example subjected to external traction forces. \textbf{(b)} 3D example subjected to external body forces. External tractions and body forces are indicated with pink arrows.}\label{fig: schematic}
\end{figure}

For the 2D case, a rectangular domain made of soft material and discretized by $8\times32$ mesh with 217 quad elements is considered. It is constrained at 4 corner nodes as shown in Figure~\ref{fig: schematic}a. It is subjected to traction forces of random magnitude, location, and direction in the region prescribed by the pink line. Body forces are ignored in this case. Table~\ref{tab:datasets} provides detailed information about datasets including the material properties and external force ranges used for the generation of 2D and 3D datasets. For the 2D case, a lower range of Y-direction force density is chosen since it doesn't contribute much to generating large deformation solutions.  

In the 3D case, a continuum toy elephant model is discretised with 6627 tetrahedron elements. It is subjected to fixed boundary conditions by constraining nodes on the bottom region of the legs. Body forces in random magnitude and direction are applied in the transverse directions (see Figure~\ref{fig: schematic}b) to generate datasets of force-displacement pairs. External tractions are ignored in this case.

\begin{table}[h!]
    \centering
    \begin{tabular}{l|R|R|L|R}
        Case& Material properties ~~
        ($E$ [Pa], $\nu$) & N.of FEM DOFs ($\mathcal{F})$ & External traction/body force density range & Dataset size (train + test)\\
        \hline
        2D domain& $100$, $0.3$ & 512 & $f_x$ = -24 to 24 $\text{N}/\text{m}$, \hspace{2em} $f_y$ = -8 to 8 $\text{N}/\text{m}$  & $7124 + 372$\\
        3D elephant& $3\!\times\!10^6$, $0.4$ & 5835 & $b_x, b_z$ = -0.35 to 0.35 $\text{N}/\text{kg}$, $b_y$=0 $\text{N}/\text{kg}$ & $7600 + 400$
    \end{tabular}
    \captionsetup{justification=centering}
    \caption{Desciption of FEM datasets}
    \label{tab:datasets}
\end{table}

\subsection{Implementation details}

As introduced in the methodology section, CNN and MAgNET frameworks belong to the family of U-Net architectures, while Perceiver IO leverages Transformer-style attention. It has to be noted that all the frameworks are robust, and do not need fine hyperparameter tuning.  

\textbf{2D case}: For the CNN U-Net, a 4-level architecture with 3 max-pooling/upsampling operations is used. At each level, two convolution layers with $3\times3$ filters are applied with 64, 128, 256, and 256 channels at respective levels. In the case of MAgNET, a 5-level graph U-Net architecture with 4 graph pooling/unpooling operations is used. At each level 2 MAg layers (with $A^2$ adjacency, refer \citep{magnet}) are applied with 8, 16, 16, 32, and 32 channels at respective levels.  For both 2D and 3D case MAgNET architectures, the seed of the first graph pooling operation is chosen by grid search, while seeds for other pooling layers are kept constant. This is done to have the maximum possible coarsened representation of the lowest-level graph. This ensures propagating boundary condition information with a minimum number of MAg operations at the lowest level. CNN U-Net is trained with a batch size of 16 for 32,000 epochs and MAgNET is trained with a batch size of 4 for 10,000 epochs. 

In the case of Perceiver IO we defined 512 inputs (standing for dofs of the example) with a total embedding size of 512 following the procedure detailed in Section~\ref{sec: inouttraining}. We also selected a total of 128 latent arrays of dimension 210 for performing cross-attention in the encoder, self attention in latent space and inputs for the decoder. For the decoder's output query array we used an index dimension of 512, which defines the size of the outputs, and a channel dimension of 210 equal to the dimension size of the latents. We used a total of 3 blocks for the latent array processing, with 2 self-attention layers per block and 2 self-attention heads per layer. Both the encoder and decoder worked with 2 cross-attention heads each. The selection of these hyper-parameters was determined in a coarse exploratory fashion, with the goal of reducing the number of network parameters while maintaining its performance. Perceiver IO is trained with a batch size of 16 for 2,64,140 epochs.   

\textbf{3D case}: For the MAgNET, 7-level graph U-Net architecture is used with 6 graph pooling/unpooling operations. Again at each level, 2 MAg layers (with $A^2$ adjacency) are applied with 6, 6, 6, 12, 12, 24, 24 channels at respective levels. The complex topology of this particular mesh demands more number graph pooling layers, this ensures propagating boundary condition information with a minimum number of MAg operations at the lowest level. On the other hand, the only change with respect to the 2D case for Perceiver IO is an increase of the input and output dimension from 512 to 5835 in both the encoder and decoder. MAgNET architecture for this case is trained for 1200 epochs with a batch size of 4, whereas Perceiver IO is trained for 32,580 epochs with a batch size of 16.

CNN U-Net and MAgNET networks are trained using the Adam optimizer \citep{adam}, whereas Perceiver IO is trained using AdamW optimizer \citep{adamw} as implemented in the original paper. CNN and MAgNET are implemented using TensorFlow \citep{tensorflow2015-whitepaper}, while Perceiver IO is implemented using PyTorch \citep{pytorch}. All the implementations in this work are performed using HPC facilities of the University of Luxembourg \citep{ULHPC}. 

\subsection{Performance on unseen examples}

Proposed DNN frameworks are trained and tested on the datasets generated as illustrated in Section~\ref{sec: data generation}. The maximum nodal displacement for the 2D case is 0.35 m and for the 3D case it is 140.04 m, i.e. we compute displacements of all the nodes for every single example, and then choose particular examples for which the maximum nodal displacement is observed. Table~\ref{tab: test metrics} summarises the performance of neural networks on the two test datasets.  It shows that all three networks are capable of predicting mechanical deformation responses with a very low error. 

To compare, we observed that CNN U-NET and Perceiver IO gave lower error metrics than MAgNET for the 2D case, which has relatively low dimensional input. As the size and complexity of the mesh increased in the 3D case, both MAgNET and Perceiver IO performed well, with Perceiver IO giving slightly better error metrics.

\bgroup
\def\arraystretch{1.5}
\begin{table}[h]
\begin{center}
 \begin{tabular}{l| c | c | c | c| c}
 Example & Framework & $M$ & $\Bar{e}$ [m] & $\sigma(e)~[\text{m}]$& $e_{\text{max}}~[\text{m}]$  \\ [0.7ex] 
 \hline
\multirow{3}{*}{2D} & CNN & \multirow{3}{*}{372} & 0.06 E-3 & 0.2 E-4 & 0.001 \\ 
& MAgNET &  & 0.17 E-3 & 0.7 E-4 &  0.021  \\
& Perceiver IO &  & 0.02 E-3 & 0.1 E-4  &  0.001 \\
 \hline
 \multirow{2}{*}{3D}& MAgNET & \multirow{2}{*}{400}  & 8.92 E-3 & 1.9 E-3 & 0.307  \\
 & Perceiver IO &  & 2.60 E-3 & 1.1 E-3  & 0.098  \\
 
\end{tabular}
\end{center}
\caption{Error metrics over the test set using the proposed NN frameworks. $M$ stands for the number of test examples, and $\Bar{e}$, $\sigma(e),e_{\text{max}}$ are error metrics defined in Section~\ref{sec: inouttraining}.} 
\label{tab: test metrics}
\end{table}

\subsection{Training and inference of DNN frameworks}

First, we compare the training convergence of proposed DNN frameworks by comparing the mean square loss plots for both 2D and 3D cases. Figure~\ref{fig: convergence}a shows that for the relatively smaller dimension inputs as in the 2D case, both CNN and Perceiver IO observed to learn more efficiently when compared to MAgNET. Figure~\ref{fig: convergence}b shows that both MAgNET and perceiver are able to learn efficiently on the complex mesh data as observed in the 3D case. However, MAgNET could learn more quickly than Perceiver IO, also MAgNET can learn efficiently even with the increased mesh complexity and input size. In case of Perceiver IO, as the size and complexity of mesh further increases, it becomes less and less robust and fails to learn efficiently. We observed that Perceiver IO failed to learn on the data with input dimension higher than $10^4$. 

A possible interpretation of this behavior is that in the case of MAgNET topological information is externally provided through the adjacency matrix. Hence MAgNET can efficiently learn by leveraging inter-dependencies between different nodal feature values in the data. On the other hand Perceiver IO implicitly learns the nodal data dependencies and as the size of the input data increases, the task to find these inter-dependencies gets more difficult. Evidence of this behavior can be seen in Figure~\ref{fig: convergence}, where it is clear that Perceiver IO is optimizing through a much more complex objective function with a higher density of local minimas. 

\begin{figure}[h!]
\begin{center}
\includegraphics[width=1\textwidth]{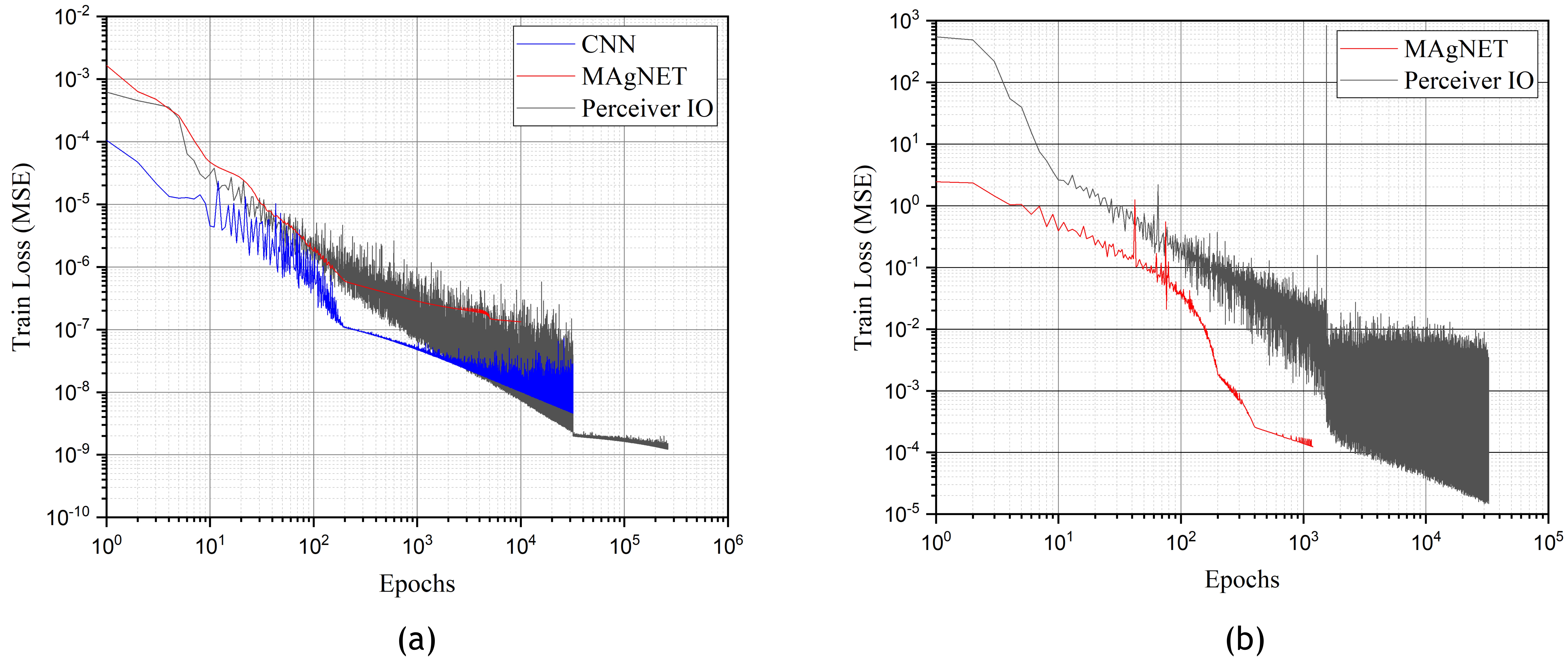}
\end{center}
\caption{Training convergence for the proposed neural network frameworks for the \textbf{(a)} 2D case \textbf{(b)} 3D case.}  \label{fig: convergence}
\end{figure}

Once trained, proposed DNN frameworks are fast at the inference stage while predicting unseen examples. Table~\ref{tab: training and test times} provides training and inference time (for a single test example) for all three frameworks, for comparison FEM solution time is also provided. In particular, Perceiver IO takes a much longer time during the training phase but is extremely fast at the inference stage. It could make predictions on both small scale (2D) and large scale (3D) inputs in almost similar time. It has to be noted that to ensure the convergence of the iterative solver, the non-linear FEM problem is solved with incremental load steps. Hence the solution time for FEM increases with the magnitude of external force. Whereas trained DNN frameworks take almost similar time at the inference stage irrespective of external force magnitudes.  

\bgroup
\def\arraystretch{1.5}
\begin{table}[h]
\begin{center}
 \begin{tabular}{l| R | R | R | R | R }
 Example & Framework & N. parameters ($\times$ E6)& Training time (hrs) & Inference time (s) & FEM solver time (s) \\ [0.7ex] 
 \hline
\multirow{3}{*}{2D} & CNN U-Net & 4.8 & 18 & 0.021 & \multirow{3}{*}{0.6}\\ 
& MAgNET & 4.5 & 132 & 0.040  &\\
& Perceiver IO & 1.9 & 521 & 0.006 &  \\
 \hline
 \multirow{2}{*}{3D}& MAgNET& 33.9 & 161 & 0.217& \multirow{2}{*}{2.5} \\
 & Perceiver IO & 4.4 &  312 &  0.006 &\\
 
\end{tabular}
\end{center}
\caption{Comparison of training and inference times for all the three networks implemented in this work.} 
\label{tab: training and test times}
\end{table}

\subsection{Qualitative analysis of individual examples}

In this section, we analyze deformations of individual examples by giving a qualitative comparison of predictions obtained using different networks. In particular, we analyze test examples with the maximum nodal displacement for 2D as well as the 3D case. In both cases, we plot nodal error contours standing for the absolute difference between the DNN prediction and the true FEM solution.

\subsubsection{2D case}
 
The analyzed example in the Figure~\ref{fig: 2d viz} stands for the maximum nodal displacement example in the 2D test dataset. The node indicated by the green dot has the maximum nodal displacement of 0.35 m. While the pink arrows represent corresponding nodal forces for the line density force applied on those four nodes. 

Figure~\ref{fig: 2d viz} shows absolute error counters of nodal displacements predicted using proposed DNN frameworks when compared with the FEM solution.  All the proposed neural network frameworks can accurately predict the deformed mesh. Percentage prediction error (when compared to the true FEM solution) for the green node is  0.03\%, 0.42\%, 0.06\% using CNN, MAgNET, and Perceiver IO network respectively.  We observed that for the small-scale structured inputs, both Perceiver IO and CNN U-Net could make better predictions when compared to MAgNET. In case of Perceiver IO, advantage likely comes from the network leveraging its capability of learning long-range correlations more accurately. This stems from the fact that Perceiver's inputs are not constrained by any topological assumption.

\begin{figure}[h!]
\begin{center}
\includegraphics[width=0.95\textwidth]{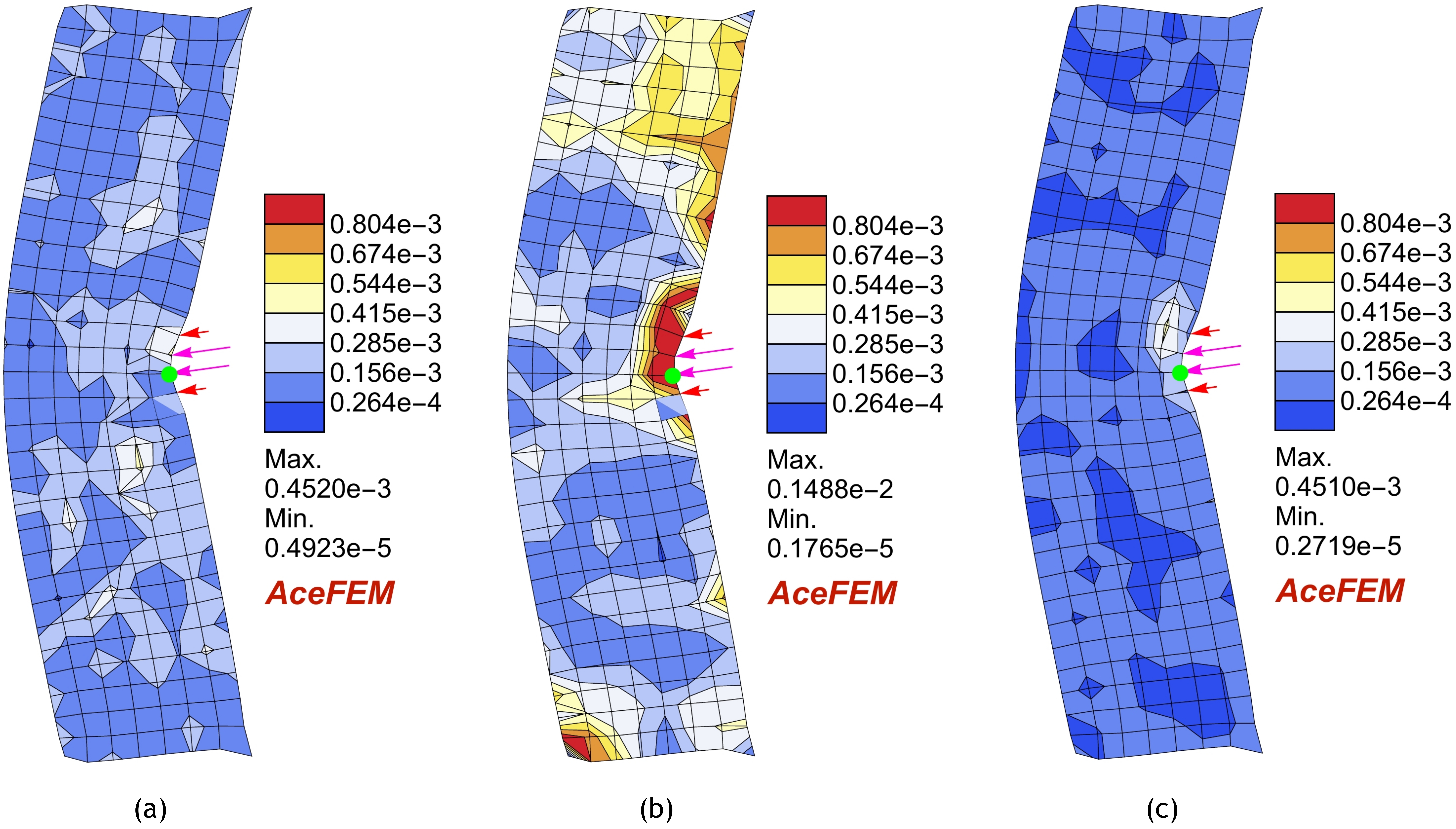}
\end{center}
\caption{Prediction error for different neural network frameworks when compared to true FEM solution, plotted on the deformed mesh (obtained using the same framework). Force with line density of (-21.6645,-2.99384) N is applied as shown with pink arrows. True displacement of the green node is 0.35m. Nodal error contours obtained \textbf{(a)} using CNN U-Net \textbf{(b)} using MAgNET \textbf{(c)} using Perceiver IO.} \label{fig: 2d viz}
\end{figure}

\subsubsection{3D case}

The 3D case is aiming to demonstrate the performance for unstructured meshes, that are commonly used when dealing with real world geometries. Tetrahedron discretisation of the continuous 3D domain is irregular and the mesh topology is complex. The fixed nodes are topologically far from the tip of the elephant trunk, thus making it challenging to communicate the boundary condition information. We show that both MAgNET and Perceiver IO can learn on such complex real-world examples efficiently.    

Again we consider the test example with maximum nodal displacement. Figure~\ref{fig: 3d viz} shows that both MAgNET (Figure~\ref{fig: 3d viz}a,\ref{fig: 3d viz}c,\ref{fig: 3d viz}e) and Perceiver IO (Figure~\ref{fig: 3d viz}b,\ref{fig: 3d viz}d,\ref{fig: 3d viz}f) solutions are able to predict non-linear mesh deformations accurately. We further analyse the absolute nodal error counters for both predictions by plotting them on the deformed meshes predicted using respective DNN frameworks. Both front (Figure~\ref{fig: 3d viz}e-\ref{fig: 3d viz}f) and side views (Figure~\ref{fig: 3d viz}c-\ref{fig: 3d viz}d) obtained using MAgNET and Perceiver IO solutions respectively indicate low prediction errors for both frameworks. The green node (at the tip of the ear) shown in (Figure~\ref{fig: 3d viz}e-\ref{fig: 3d viz}f) has a maximum nodal displacement of 140.04 m for this example. The percentage prediction error when compared to the true FEM solution for this green node is 0.03\%, and 0.02\% using MAgNET and Perceiver IO networks respectively. Both MAgNET and Perceiver IO are observed to make efficient predictions, with Perceiver IO giving relatively low nodal errors for this demanding case. Also, owing to the lesser number of trainable parameters, Perceiver IO is much faster at the inference stage.

\begin{figure}[h!]
\begin{center}
\includegraphics[width=0.9\textwidth]{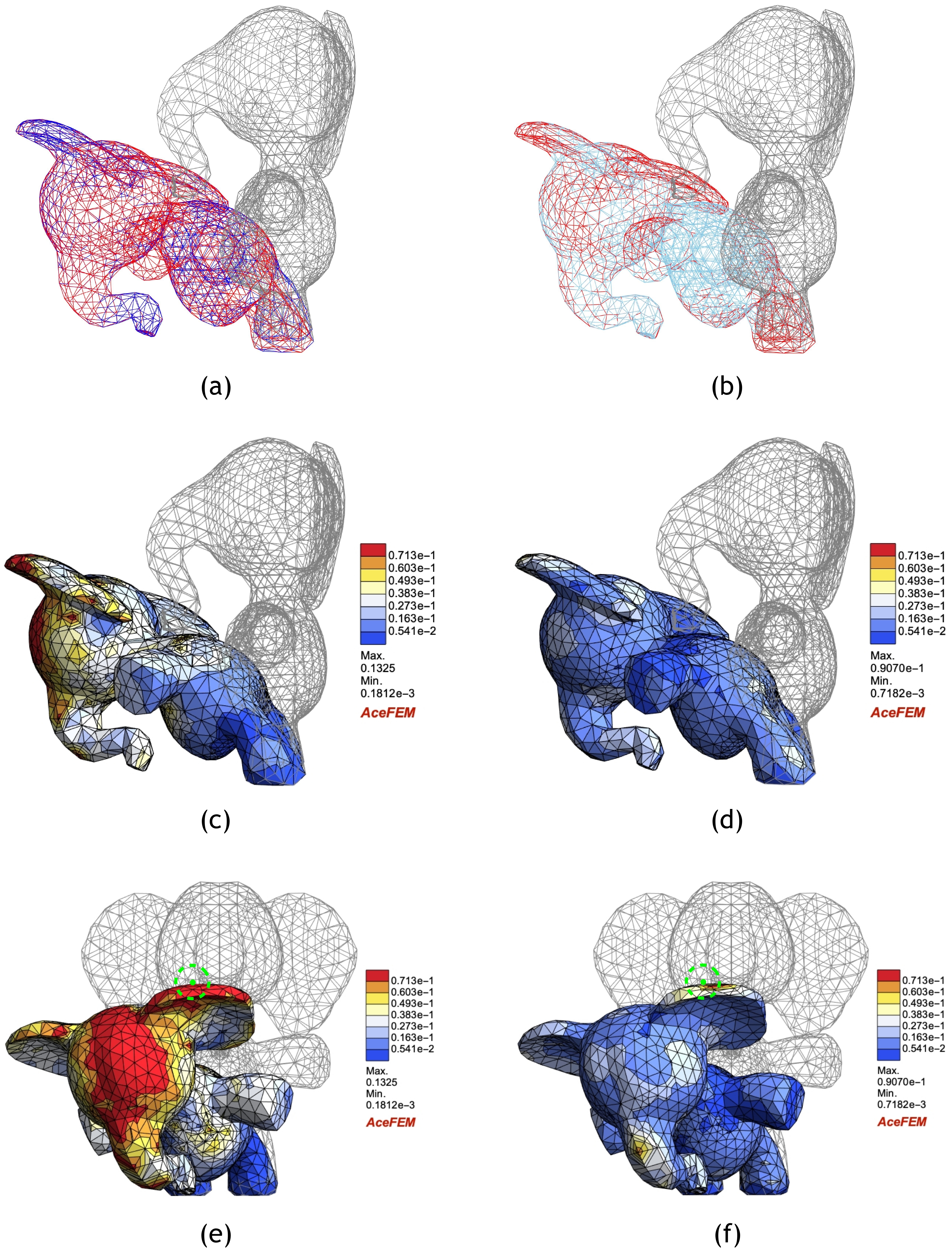}
\end{center}
\caption{Deformation of elephant mesh subjected to external body force density (0.34, 0.0, 0.35) $\text{N}/\text{kg}$. First column represents MAgNET solutions while the second column represents Perceiver IO solutions. \textbf{(a)\&(b)} Deformed meshes using MAgNET~(dark blue) and Perceiver IO~(sky blue) respectively, for comparison FEM mesh is presented in red. The rest position is indicated with gray mesh. \textbf{(c)\&(d)} Side view of nodal error contours when compared to the FEM solution, plotted on the deformed meshes for MAgNET and Perceiver IO solution respectively. \textbf{(e)\&(f)} Front view of nodal error contours for MAgNET and Perceiver IO respectively. The true displacement of the green node is 140.04 m.}
\label{fig: 3d viz}
\end{figure}

\section{Conclusion}
\label{sec: conclusion}

In this work, we demonstrated the capabilities of three promising deep neural network (DNN) frameworks for accurate and fast predictions of non-linear deformations of solid bodies. We compared their performance on two benchmark examples, in which data was generated by the finite element method. Although we only tested the frameworks for the Noe-Hoohean material model, they are compatible with more general hyperelastic models, such as  Mooney–Rivlin or Ogden models. As such, they promise to be used as surrogate models for non-linear computational models in mechanics.

The comparison included two very recent DNN frameworks, MAgNET and Perceiver~IO, that are naturally able to work with arbitrarily structured data at inputs/outputs, including complex finite element meshes that originate from real-world applications. The third compared framework, CNN U-Net, could only operate on grid inputs/outputs, and we suggested possible remedies to extend it to work with arbitrary unstructured meshes. When looking at prediction capabilities, especially interesting are the capabilities of the Perceiver~IO network, which demonstrated to give better predictions with a lesser number of parameters, as compared to MAgNET and CNN U-Net. Additionally, the use of Perceiver~IO creates a direct link to rapidly advancing research in ML and AI communities, which promises further advancements. 

MAgNET and Perceiver IO are designed to be flexible in terms of the input and output structures, allowing them to potentially be applied to a wide range of problems. One possible application for these types of neural networks is in \emph{ab initio} multi-scale modeling, which is also pursued in our team, see~\cite{Hauseux2020}. These methods could be used to accelerate computationally expensive accurate simulations of large atomic systems by helping to connect atomic-level simulations with the macroscopic continuum description of materials. As such, these neural networks could lead to significant strides in the field of materials science.

One of the first possible future extensions of the presented frameworks would be to incorporate the physics-informed neural network paradigm. This can be easily achieved by incorporating relevant physical laws in the optimization objective of the training procedure. Such extension can further increase the accuracy of predictions and accelerate the training procedure. Another possible extension is to consider a much wider class of phenomena and models, including buckling instabilities and more general history/time-dependent phenomena (visco-elasticity, dynamics, plasticity, etc.), which would allow tackling more challenging problems in solid mechanics, see e.g., \citep{soumi}. Going beyond mechanics, these approaches can be also adopted for a much wider range of engineering and scientific applications.



\section*{Conflict of Interest Statement}

The authors declare that the research was conducted in the absence of any commercial or financial relationships that could be construed as a potential conflict of interest.

\section*{Author Contributions}

SD conceptualized and created the core structure of the manuscript with the help of JL. SD and RIS carried out all the simulations presented in this work. SD wrote the main part and assembled all of the manuscript, RIS contributed to multiple sections. JL and SPAB revised the text in detail. All authors read, discussed, and approved the final version.


\section*{Acknowledgments}
\begin{wrapfigure}{l}{0.34\textwidth}
    \includegraphics[width=0.3\textwidth]{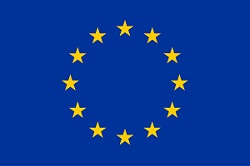}
\end{wrapfigure}

This project has received funding from the European Union’s Horizon 2020 research and innovation programme under the Marie Sklodowska-Curie grant agreement No. 764644. 
Jakub Lengiewicz would like to acknowledge the support from EU Horizon 2020 Marie Sklodowska Curie Individual Fellowship \emph{MOrPhEM} under Grant 800150.
This paper only contains the author's views and the Research Executive Agency and the Commission are not responsible for any use that may be made of the information it contains. 

St\'ephane Bordas, Jakub Lengiewicz and Ra\'ul I. Sosa are grateful for the support of the Fonds National de la Recherche Luxembourg FNR grant QuaC C20/MS/14782078. St\'ephane Bordas received funding from the European Union's Horizon 2020 research and innovation programme under grant agreement No 811099 TWINNING Project DRIVEN for the University
of Luxembourg.


\section*{Data Availability Statement}
The datasets generated for this study can be found at [\href{https://doi.org/10.5281/zenodo.7585319}{https://doi.org/10.5281/zenodo.7585319}] zenodo repository, and source codes are made available in the \href{https://github.com/saurabhdeshpande93/convolution-aggregation-attention.git}{convolution-aggregation-attention} GitHub repository. 

\bibliographystyle{Frontiers-Harvard}

\bibliography{frontiersbib}

\end{document}